# Linguists should learn to love speech-based deep learning models

Marianne de Heer Kloots[1], Paul Boersma[2], Willem Zuidema[1]

[1] Institute for Logic, Language and Computation, University of Amsterdam
[2] Amsterdam Center for Language and Communication, University of Amsterdam

Correspondence: m.l.s.deheerkloots@uva.nl

**Abstract**
Futrell and Mahowald present a useful framework bridging technology-oriented deep learning systems and explanation-oriented linguistic theories. Unfortunately, the target article's focus on generative text-based LLMs fundamentally limits fruitful interactions with linguistics, as many interesting questions on human language fall outside what is captured by written text. We argue that audio-based deep learning models can and should play a crucial role.

**Main text**
To integratively study a human spoken language, linguists investigate how speakers map their communicative intent all the way to their articulatory gestures (language production), as well as how listeners map incoming auditory signals to their interpretation of a speaker's intent (language comprehension). An inclusive linguistics is therefore majorly concerned with structural relations between physical signals and linguistic content — relations typically studied by the fields of *phonetics* (audition, acoustics, articulation) and *phonology* (sound structure). As these fields are largely irrelevant to text-based LLMs, F&M don't address them. This jeopardizes their own endeavour.

**The trouble with text.** For language comprehension, LLMs operate on text that has already discretized the continuous auditory speech stream into things that look like words, morphemes, syllables, and/or phonological segments (vowels, consonants), depending on the language. Conversely, crucial phonological aspects of spoken language like prosody and intonation aren't usually represented in text. As a result, ambiguities differ between text and speech: English text but not speech distinguishes between *sun* and *son*, while English speech but not text disambiguates the polar vs. alternative readings of *Do you like coffee or tea?* In general, text-based models end up solving fundamentally different problems than human spoken language users, as both cognitive scientists and language technologists have noted before (Dupoux, 2018; Chrupała, 2023).

The structures of natural languages reflect overwhelmingly the properties of speech (or signing), rather than those of text. Merely enriching text-based LLMs with speech capacities through acoustic tokens or separate speech recognition/synthesis components (Arora et al., 2025) cannot resolve the deep trouble caused by textual bottlenecks in linguistic modelling. Researchers interested in modelling linguistic structure should therefore not settle for LLMs, but rather learn to love models designed to capture the more natural form of human spoken language: the speech signal itself.

**Linguistic structure in models of speech.** Many current speech-technological applications employ self-supervised *speech foundation models*, i.e. deep learning architectures trained to represent audio signals on the basis of unlabelled speech recordings. Parallel to work on the interpretability of text-based LLMs (as covered by F&M), *linguistic interpretability* studies in the speech domain investigate to what extent speech-based models learn to capture any higher-level patterns that make up spoken language. Method-wise, most of these studies use *representational probes* to examine what linguistic information is represented in speech models' internal states, with insightful results obtained for the encoding of linguistic units like phonemes (Alishahi, Barking, & Chrupała, 2016; Martin et al., 2023) and words (Pasad et al., 2024), as well as morphophonological (Gauthier et al., 2025) and suprasegmental patterns (Shen et al., 2024); other studies use *behavioural (minimal pair) tests* to ask e.g. whether models can distinguish words from pseudowords (Lavechin et al., 2023) or prosodically natural vs. unnatural pauses (de Seyssel et al., 2023). With these types of methods, speech models can be used as "psycholinguistic participants", mimicking human speech perception experiments and their results, e.g. about perceptual similarity (Millet & Dunbar, 2022) and phonetic categorization (de Heer Kloots & Zuidema, 2024).

Linguists can already draw an important theoretical insight from these findings: speech-only models can indeed learn relevant patterns of linguistic structure, without the pre-existing symbolic categories assumed by earlier connectionist work (McClelland & Elman, 1986; Smolensky, 1990, 1999).

**Inductive biases from domain-general perception.** Further investigations of what deep learning models learn from audio can potentially inform what mechanisms drive language-relevant perceptual learning in humans. Here, we agree with F&M that deep learning models can helpfully serve as proof-of-concepts for ideas that were previously hard to formalize. At least part of the human perceptual system involved in speech processing is also involved in processing a wide variety of other auditory signals. Deep learning models pre-trained on music and/or environmental sounds show more human-like behaviour than models trained on speech sounds alone, in detecting algebraic patterns in tone or syllable sequences (Orhan, Boubenec, & King, 2025), and in displaying native language effects when processing foreign speech sounds (Poli et al., 2024). Hence, it seems that some inductive biases relevant to the encoding of language-like structures can result from (evolutionarily) optimizing the perceptual system for representing sounds other than speech, perhaps more so than considered before (Aslin & Pisoni, 1980; Soderstrom, Mathis, & Smolensky, 2006).

**Inductive biases from bidirectional processing.** One property of much cognitively grounded neural modelling, that is not shared with current technological speech or text machines, is that connection weights are symmetric (Kohonen, 1982; Hopfield, 1982; McMurray et al., 2009; Salakhutdinov & Hinton, 2009). In linguistics this corresponds to a specific inductive bias, namely that language employs *bidirectional processing*, i.e. language comprehension and language production utilize the same knowledge.

In phonetics and phonology, neural models with bidirectional connection weights straightforwardly predict the diachronic evolution of auditory dispersion in phoneme inventories (Boersma, Benders & Seinhorst, 2020). Unpublished simulations show that such communicative-success-optimizing effects readily transfer to other linguistic subdomains: with bidirectional connection weights, the maxims of Grice (1975) emerge in pragmatics, as do anti-synonymity effects in semantics, morphology and syntax (Boersma, 2009).

A drawback of small neural models is that they work exclusively on *toy problems*, rather than *whole languages*. Hopefully evidence from toy models can nevertheless inspire speech technologists in architecture design. Most findings cited above were obtained with speech models trained on developmentally realistic amounts of speech (< 1000 hours); beyond improving data efficiency, exploring a variety of inductive biases is primarily crucial in developing more human-like systems with relevance for linguistic theory.

**Take-home message.** We agree that neural network models developed for technological applications can provide insights into human language and cognition. The early connectionist work that F&M describe as preceding the LLM era included efforts to take the continuous speech signal seriously (Elman & Zipser, 1988; Norris, 1994), and we argue that current linguistic investigations with deep learning models can and should do the same. Once the bottleneck of text can be replaced, we stand a better chance of modelling more human-like linguistic processes, as well as handling the vast majority of local and regional language varieties that are rarely or never written down.